\documentclass[final,5p,times,twocolumn]{elsarticle}
\usepackage{math}
\usepackage{amsmath}
\usepackage{amssymb}
\usepackage[latin1]{inputenc} 
\usepackage[T1]{fontenc} 

\journal{Computer Vision and Image Understanding, vol. 93, pp 175--194, 2004}

\begin{document}
\begin{frontmatter}

\title{Image Matching with Scale Adjustment}
\author{Yves Dufournaud, Cordelia Schmid, and Radu Horaud}
\address{INRIA Grenoble Rh\^{o}ne-Alpes, Montbonnot Saint-Martin, France}

\begin{abstract}
In this paper we address the problem of matching two images with two
different resolutions: a high-resolution image and a low-resolution
one. The difference in resolution between the two images is not known
and without loss of generality one of the images is assumed to be the
high-resolution one. On the premise that changes in resolution act
as a smoothing equivalent to changes in scale, a scale-space
representation of the   high-resolution image is produced. Hence the
one-to-one classical image matching paradigm becomes one-to-many
because the low-resolution  image is compared with all the scale-space
representations of the high-resolution one. Key to the success of such
a process is the proper representation of the features to be matched
in scale-space. We  show how to represent and extract interest points at variable
scales and we devise a method allowing the comparison of two images at
two different  resolutions. The method comprises the use of
photometric- and rotation-invariant descriptors, a geometric model
mapping the high-resolution image onto a low-resolution image region,
and an image matching strategy based on local constraints and on the robust estimation of this
geometric model. Extensive experiments show that our matching method
can be used for scale changes up to a factor of 6.
\end{abstract}


\end{frontmatter}

\section{Introduction}

The problem of matching two images has been an active topic of
research in computer vision for the last two decades. The vast
majority of existing methods consider two views of the same scene
where the viewpoints differ by small offsets in position, orientation
and viewing parameters such as focal length. Under such conditions,
the image features associated with the two views have comparative resolutions
and hence they encapsulate scene features which appear in the two
images at approximatively the same
scale. In this paper we address a somehow different problem that
has received little attention in the past. We consider the problem of
matching two images with very different resolutions.

Obviously, the resolution with which a 3-D object is observed in an
image mainly depends on two factors: the distance $d$ from camera to
object and the focal length $f$ associated with the camera lens. Image 
resolution increases with $f$ and decreases with $d$. Therefore,
$r=f/d$ is a good, first-order approximation, measure of image
resolution. We are interested in developing matching techniques which
take as input an image pair whose resolutions are quite different,
$r_1 << r_2$. In practice we will describe an image-matching technique 
which takes as input a low-resolution image (image~\#1) and a
high-resolution one (image~\#2). It will be shown
that, using the approach advocated below, it is possible to match two
images satisfying $r_2/r_1=6$.

\begin{figure}[t]
\centerline{
\includegraphics[width=0.48\columnwidth]{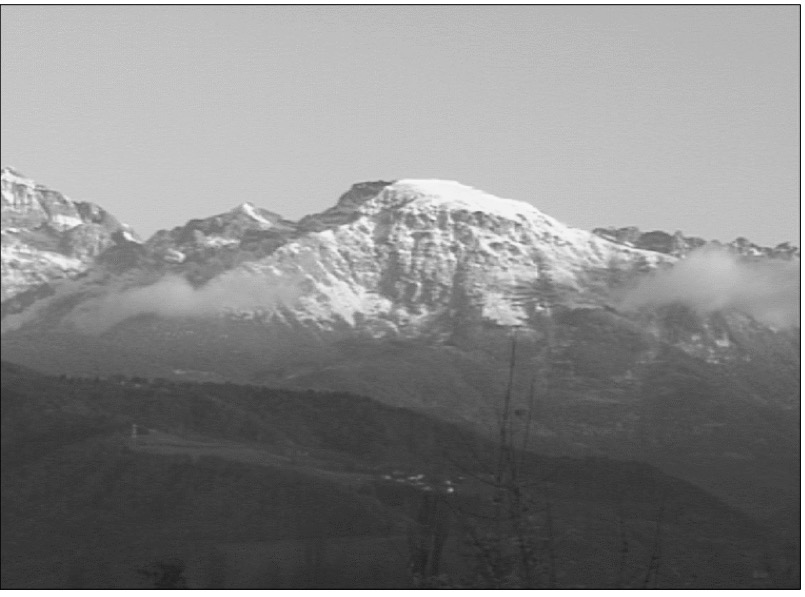} 
\hfill
\includegraphics[width=0.48\columnwidth]{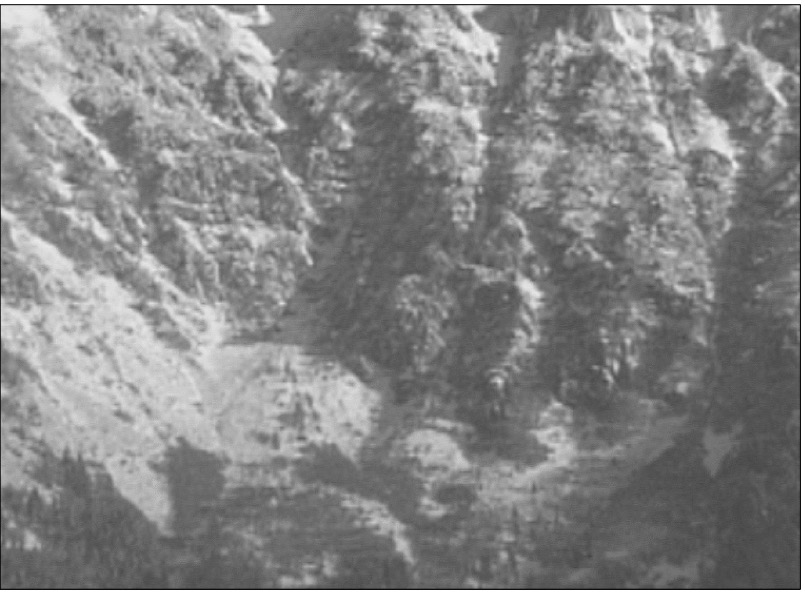}}
\caption{An example of an image pair with different resolutions:
low-resolution (left) and high-resolution (right).}
\label{fig:Grand-Colon-image-pair}
\end{figure}

As an example we consider the image pair in
Figure~\ref{fig:Grand-Colon-image-pair}. Both images were
taken with a camera placed at 11 kilometers (6.9 miles) away from the
top of the  mountain. For the first image (left) we used a focal
length equal to 12mm while for the second one (right) we used a focal
length equal to 72mm. Notice that the high-resolution image
corresponds to a small region of the low-resolution one and it is
quite difficult to find the exact position and size of this
region. Moreover, the low-resolution image (left) covers in practice a 
wide range of resolutions because scene objects appear at various
depths values.

Therefore, the search space associated with the feature-to-feature
matching of two such images is larger and more complex than the one
associated with the classical stereo matching paradigm. The classical
approach to image matching proceeds as follows: (i)~extract interesting point-features from
each image, (ii)~match them based on cross-correlation, (iii)~compute the
epipolar geometry through the robust estimation of the fundamental
matrix, and (iv)~establish many other matches once this matrix is known. 
For a number of reasons, this method cannot be applied
to the problem at hand:
\begin{enumerate}
\item Point-feature extraction and matching are resolution-dependent
processes.
\item The high-resolution image corresponds to a small region of the
low-resolution one and hence the latter contains many features
which do not have a match in the former.
\item It may be difficult to estimate the epipolar geometry because
there is not enough depth associated with both the high resolution
image and its associated small area of the low-resolution image.
\end{enumerate}

The solution suggested in this paper consists of considering a
scale-space representation of the high-resolution image and of
matching the low-resolution image against the scale-space descriptions
of the high-resolution one. A scale-space 
representation may be obtained by smoothing an image with Gaussian
kernels of increasing standard deviations. Therefore, the
high-resolution image will be described by a discrete set of images at 
various scales. On the premise that decreasing the resolution can be
modeled as image smoothing which is equivalent to a scale change, the
one-to-one image matching problem at hand becomes a one-to-many image
matching problem \cite{DufournaudSchmidHoraud00}.

In this paper we describe such a matching method. Key to its success
are the following characteristics:
\begin{itemize}
\item  The scale-space representation of image point features (or interest points) together with
their associated descriptors;
\item A geometric model describing the mapping from the
high-resolution image to a region of the low-resolution one.
\item An image-matching strategy combining point-to-point
assignments with a robust estimation of the geometric mapping between
image regions.
\end{itemize}

Several authors addressed the problem of matching two images gathered
from two very different viewpoints
\cite{georgis98b,pritchett98a,tell00a,tuytelaars99a} 
but they did not consider a large change in resolution. The use of
scale-space in conjunction with stereo matching has been restricted to
hierarchical matching: correspondences obtained at low resolution
constrain the search space 
at higher resolutions \cite{glazer83a,quam87a,lew99a}. Scale-space
properties are thoroughly studied in \cite{lindeberg94a} and the
same author attempted to characterize the best scale at which an image 
feature should be represented \cite{lindeberg98a}. 
A similar idea is presented in \cite{lowe99a} to detect stable points
in scale space.

Our work is closely related to \cite{hansen99a} who attempts to 
match two images of the same object gathered with two different zoom
settings. Point-to-point correspondences are characterized in scale
space by correlation traces. The method is able to recover the scale 
factor for which two image points are the most similar but it cannot
deal with camera motions.

Local descriptors that are invariant with respect to affine
grey value changes, image rotations, and image translations were
studied theoretically in \cite{konderink87a} and were used in the
context of image matching in \cite{schmid97a}. These descriptors are
based on convolutions with Gaussian kernels and their derivatives.
Therefore they are consistent with scale-space representations. They are
best applied at image locations found by interest points and a recent study showed that the
Harris corner detector \cite{harris88a} is the most reliable interest
point detector
\cite{schmid00a}. However, these local descriptors are not scale-invariant and, in spite
of good theoretical models for scale-space invariants
\cite{konderink84b,lindeberg94a}, it is more judicious, from a
practical point of view, to compute local descriptors at various scales
in a discrete scale-space \cite{schmid97a}. 

The main contributions of this paper are the followings. We thoroughly
study the behaviour of the Harris interest point detector under a
similarity transformation. This detector comprises convolutions with
two Gaussian kernels, one for weigthing and one for computing
grey-level derivatives. We show under which conditions the detector is 
invariant to rotations and translations in the image plane. Based on
this we derive a scale-space representation of interest points. This
representation allows to match points from images at very different
resolutions, which has never been performed in the past -- up to a
factor of 6. In order to match points we describe a way to represent
local collections of points and we seek similarities between such
local collections at different scales. Finally a one-to-many image
matching technique (with scale adjustment) is described. Many examples 
with various scenes, camera configurations and settings illustrate the 
method both quantitatively and qualitatively.

\paragraph{Paper organization} The remainder of this paper is
organized as follows. Section~2 briefly outlines the geometric model
associated with the image pair. Section~3 suggests a framework for
adapting the detection of interest points to scale changes, image
rotations, and image translations. Section~4
describes the high-resolution to low-resolution matching and section~5
presents experimental results.

\section{Geometric modeling}

One of the key observations enabling the matching of two images at two
different resolutions is that the high-resolution image corresponds to 
a small region of the low-resolution one. Without loss of generality,
it may be assumed that the high-resolution image has homogeneous
resolution because the observed 3-D features are, approximatively at
the same distance. Clearly this is not the case for the low resolution 
image which contains various features at various ranges. The matching
task therefore consists in finding a small region in the low
resolution image that can be assigned to the whole high resolution one.

One reasonable
assumption is to consider that the mapping between the high resolution 
image and the corresponding low-resolution region is a plane projective 
transformation, i.e., the scene corresponding to this region is
planar. Such a homography may well be represented by a 3$\times$3
homogeneous full rank matrix $\Hmat$. Let $\mvect$ be a point in the 
high-resolution image $I$ and $\mvect'$ be a point in the low-resolution
image $I'$. One can characterize a region in the
low-resolution image such that the points $m'\in{\cal R}$ within this region verify:
\begin{equation}
\mvect' \prj \Hmat \mvect
\label{eq:plane-homography-inliers}
\end{equation}
Similarly, points outside this region do not
verify this equation. In general, image descriptors which are
invariant to such a general plane-to-plane projective transformation
are difficult to compute and therefore it is difficult to properly
select potential candidate points satisfying
eq.~(\ref{eq:plane-homography-inliers}). 

We can further simplify the geometric model and consider a restricted
class of homographies, namely a rotation about the optical axis by an
angle~$\theta$, a
translation in the image plane by a vector $(a,b)$, and a {\em similitude} factor~
$h$:
\begin{equation}
\mvect'  =\left[ \begin{array}{ccc}
    h\cos\theta & -h\sin\theta & a \\
    h\sin\theta & h\cos\theta & b \\
    0           & 0           & 1 
  \end{array} \right]
\mvect
\label{eq:plane-affinity}
\end{equation}
Notice that the projective equality in
eq.~(\ref{eq:plane-homography-inliers}) is replaced by an equality. 
In practice it will be useful to replace
the 3-vectors $\mvect$ and $\mvect'$ used above by 2-vectors $\xvect$
and $\xvect'$ such that:
\[
\mvect' = \left( \begin{array}{c} u' \\ v' \\ 1 \end{array} \right) =
\left( \begin{array}{c} \xvect' \\ 1 \end{array} \right)
\mbox{ and }
\mvect = \left( \begin{array}{c} u \\ v \\ 1 \end{array} \right) =
\left( \begin{array}{c} \xvect \\ 1 \end{array} \right)
\]
With this notation, eq.~(\ref{eq:plane-affinity}) becomes:
\begin{equation}
\xvect' = h\Rmat \xvect + \tvect
\label{eq:plane-similarity}
\end{equation}
where $\Rmat$ is the 2$\times$2 rotation matrix and $\tvect$ is the
translation vector.

In order to match two images which differ by such a geometric
transformation, one has to define a measure of
similarity. One possibility is to use 
correlation. In this case, the similarity between $\xvect\in I$ and
$\xvect'\in I'$  can be written as:
\[
\sum_{\Delta\pvect}
\left[ I'(\xvect' - \Delta\pvect') - I(\xvect - \Delta\pvect) \right]^2
\]
where $\Delta\pvect$ and $\Delta\pvect'$ are shift vectors. With the substitution
for $\xvect'$ above, i.e., eq.~(\ref{eq:plane-similarity}) and with
$\Delta\pvect'=h\Rmat\Delta\pvect$ 
we obtain:
\begin{equation}
\sum_{\Delta\pvect}
\left[I'(h\Rmat(\xvect - \Delta\pvect)+\tvect)  - I(\xvect -
\Delta\pvect) \right]^2
\label{eq:similarity-measure}
\end{equation}

Therefore, one must find a scale factor $h$, a rotation matrix
$\Rmat$, and a translation vector $\tvect$
for which the expression above is minimized. The search space associated
with such a technique is very large and the associated non-linear
minimization procedure has to deal with a four-parameter cost function 
\cite{gruen85a}.

\section{Interest point detection for image matching}

Alternatively, one may use interest points.
Ideally, one would like to characterize such image points by descriptors
invariant to image rotation,  translation and scale. Unfortunately,
scale-invariant image descriptors are hard to 
compute in practice. Therefore, the matching strategy will build a
discrete scale-space for the high-resolution image thus
by-passing the scale-invariance problem. 
The image matching problem at hand then becomes a one-to-many image
matching technique. 

The steps for image-to-image matching are:
\begin{itemize}
\item[(i)] extract
sets of interest points from the two images, $(\xvect_1,\ldots,\xvect_M)$ and 
$(\xvect'_1,\ldots,\xvect'_N)$, 
\item[(ii)] characterize these points
such that point-to-point comparisons are made possible, and 
\item[(iii)] determine
the largest set of such correspondences compatible with a similarity
between the high-resolution image and a low-resolution region. 
\end{itemize}

The one-to-many matching algorithm uses this procedure and the image pair
with the highest matching score determines the appropriate
scale for matching and allows to estimate the scale change. The
advantage of this approach mainly resides in step (iii) above. Two
point-to-point correspondences are sufficient to estimate the
similarity parameters described in eq.~(\ref{eq:plane-affinity}) (four such
correspondences are necessary for a full homography) and therefore the 
largest set of point correspondences is found by an efficient robust
estimator.

\subsection{Interest point detection under similarity}

We use the interest point detector proposed in~\cite{harris88a}. 
This operator was studied experimentally and it was shown to be
robust to image rotations, translations and
illumination changes~\cite{schmid00a}. However, the Harris point
detector is not invariant to
changes in scale. In this section and in the next section we derive an exact formula for
analyzing the behaviour of this interest-point detector over changes
in scale, rotation, and translation.

We consider as before two images $I(\xvect)$ and $I'(\xvect')$ with
$\xvect=(u,v)\tp$ and $\xvect'=(u',v')\tp$. 

An interest point is detected in image $I$ (or in image $I'$) as follows: 
\begin{enumerate}
\item Compute the image derivatives in the $u$ and $v$ directions,
$I_u$, and $I_v$. These computations are carried out by convolution
with the differential of a Gaussian kernel of standard deviation
$\sigma$: 
\begin{eqnarray*}
I_u(\xvect, \sigma) &=& I(\xvect) \star G_u(\xvect,\sigma)\\
I_v(\xvect, \sigma) &=& I(\xvect) \star G_v(\xvect,\sigma)\\
I_uI_v(\xvect, \sigma) &=& I_u(\xvect, \sigma) \, I_v(\xvect, \sigma)
\end{eqnarray*}

\item Form the auto-correlation matrix  $\Mmat(\xvect, \sigma, \tilde{\sigma})$.
This matrix
sums up derivatives in a window around a point
$\xvect$ with a Gaussian kernel $G(\xvect,\tilde{\sigma})$ being used for weighting: 
\begin{multline}
\label{eq:Harris}
\Mmat(\xvect, \sigma, \tilde{\sigma}) =  \\
\left[ \begin{array}{ll} 
G(\xvect,\tilde{\sigma}) \star I_u^2(\xvect, \sigma) & G(\xvect,\tilde{\sigma})
\star I_u I_v(\xvect, \sigma) \\
G(\xvect,\tilde{\sigma}) \star I_u I_v(\xvect, \sigma) & G(\xvect,\tilde{\sigma})
\star I_v^2(\xvect, \sigma)
\end{array} \right] 
\end{multline}

\item $\xvect$ is an interest point if the matrix $\Mmat$ has two
significant eigenvalues, that is, if the determinant and trace of
this matrix verify a measure of ``cornerness'':
\begin{equation}
{\cal C}(\xvect) = \det(\Mmat(\xvect)) - \alpha \, \trace(\Mmat(\xvect)) ^2
\label{eq:corner-detection}
\end{equation}
where $\alpha$ is a fixed parameter. An interest point is detected at
image location $\xvect$ if ${\cal C}(\xvect)>t$, where $t$ is a threshold.
\end{enumerate}

In order to study the behaviour of this operator to changes in scale,
rotation, and translation, let us introduce the following notation:
\begin{equation}
\Mmat(\xvect, \sigma, \tilde{\sigma}) = G(\xvect,\tilde{\sigma}) \star
\Qmat(\xvect, \sigma)
\label{eq:C=GQ}
\end{equation}
with:
\begin{equation}
\Qmat(\xvect, \sigma) = 
\left[ \begin{array}{ll} 
I_u^2(\xvect, \sigma) & I_u I_v(\xvect, \sigma) \\
I_u I_v(\xvect, \sigma) & I_v^2(\xvect, \sigma)
\end{array} \right]
= \left( \begin{array}{c} I_u \\ I_v \end{array} \right)
\left( \begin{array}{cc} I_u & I_v \end{array} \right)
\label{eq:Qdefinition}
\end{equation}

Under the assumption that the two images are properly normalized, the
condition that must be satisfied is the equality of the two image intensities at
two pixels:
\begin{equation}
I'(\xvect') = I(\xvect)
\label{eq:I=I'}
\end{equation}
This allows us to build a relationship that must hold between the
autocorrelation matrices associated with two matching points in the
two images and between
cornerness measurements $\cal{C}$ and $\cal{C}'$ associated with the
autocorrelation matrix. The following proposition establishes these relationships:
\begin{proposition}\label{prop:similarity}
The auto-correlation matrices at locations $\xvect$ (in image $I$) and 
$\xvect'$ (in image $I'$) are related by the following formula provided that the 
standard deviation of the smoothing Gaussian kernels are choosen such
that $\tilde{\sigma}'=h\tilde{\sigma}$:
\begin{equation}
\Mmat'(\xvect',\sigma', \tilde{\sigma}') = \frac{1}{h^2}\Rmat
\Mmat(\xvect, \sigma, \tilde{\sigma}) \Rmat\tp 
\label{eq:CandC'}
\end{equation}
The equivalent relationship between the two cornerness measurements is 
given by the formula:
\begin{equation}
{\cal C}' = \frac{1}{h^4} {\cal C}
\label{eq:cornerness-rel}
\end{equation}
\end{proposition}
\proof
Noticing that the trace and determinant of a matrix are invariant with 
respect to a similarity transformation, i.e.,
$\Amat\rightarrow\Bmat\inverse\Amat\Bmat$, it is straightforward to
derive eq.~(\ref{eq:cornerness-rel}) from eqs.~(\ref{eq:CandC'}) and
(\ref{eq:corner-detection}).

In order to show that eq.~(\ref{eq:CandC'}) holds, let us derive both 
sides of eq.~(\ref{eq:I=I'}) with
respect to $u$ and $v$:

\[
 \left( \begin{array}{c} I_u \\ I_v \end{array} \right)
= \left( \begin{array}{c} \frac{\partial I}{\partial u} \\ \\
\frac{\partial I}{\partial v} \end{array} \right) 
=\left( \begin{array}{c} \frac{\partial I'}{\partial u} \\ \\
\frac{\partial I'}{\partial v} \end{array} \right)
= \left( \begin{array}{c} I'_u \\ I'_v \end{array} \right)
\]
The relationship between the pixel coordinates $\xvect'=h\Rmat\xvect + 
\tvect$ combined with the chain rule of derivation allows us to otbain:

\[
\left( \begin{array}{c} I'_u \\ I'_v \end{array} \right)
= \left[ \begin{array}{cc} \frac{du'}{du} & \frac{dv'}{du} \\ & \\
			   \frac{du'}{dv} & \frac{dv'}{dv}
\end{array} \right]
\left( \begin{array}{c} I'_{u'} \\ I'_{v'} \end{array} \right)
= h \Rmat\tp
\left( \begin{array}{c} I'_{u'} \\ I'_{v'} \end{array} \right)
\]

The formulae above allow us to express a relationship between the
quadratic forms $\Qmat(\xvect,\sigma)$ and $\Qmat'(\xvect',\sigma')$,
i.e., eq.~(\ref{eq:Qdefinition}):
\begin{equation}
\Qmat(\xvect,\sigma) = h^2\Rmat\tp \Qmat'(\xvect',\sigma') \Rmat
\label{eq:QandQ'}
\end{equation}

Finally, using the properties of convolution applied to
eq.~(\ref{eq:C=GQ}) we obtain the formula given by eq.~(\ref{eq:CandC'})
(see appendix~\ref{appendix:Harris-over-scale} for a formal derivation).
\endproof

\subsection{Interest point detection and scale-space}

We consider now the scale-space associated with the high-resolution
image $I$.
The scale-space is obtained by convolving the initial image
with a Gaussian kernel whose standard deviation is increasing
monotonically, say $s\sigma$ with $s>1$. At scale $s$ we have the
following image derivatives that allow the estimation of interest
points:
\begin{eqnarray*}
I_u(\xvect, s\sigma) &=& I(\xvect) \star G_u(\xvect,s\sigma)\\
I_v(\xvect, s\sigma) &=& I(\xvect) \star G_v(\xvect,s\sigma)
\end{eqnarray*}

If the task consists of matching a
high-resolution image $I$ with a low-resolution one $I'$, it is crucial to
select the scale of $I$ at which this matching has to be
performed. 
The scale parameter $s$ must ``absorb'' the similarity factor $h$ such 
that interest points that are detected in image $I$ at scale $s$ best correspond to 
interest points detected in image $I'$. Since the resolution of $I$
decreases with increasing $s$ one needs to set:
\[
s = \frac{1}{h}
\]

The scale-space interest-point detector is then defined as
follows. From eq.~(\ref{eq:CandC'}) and with the relationship between
$s$ and $h$ we obtain the autocorrelation matrix:
\begin{multline}
\label{eq:Harris-scale}
 \Mmat_s(\xvect, s\sigma, s\tilde{\sigma})
= \\
s^2 G(\xvect, s\tilde{\sigma}) 
  \star 
\left[ \begin{array}{cc} 
I_u^2(\xvect, s\sigma) & I_u I_v(\xvect, s\sigma) \\
I_u I_v(\xvect, s\sigma) & I_v^2(\xvect, s\sigma)
\end{array} \right] 
\end{multline}
The cornerness measure becomes:
\[
{\cal C}_s(\xvect) = s^4 \left(
\det(\Mmat_s(\xvect)) - \alpha \, \trace(\Mmat_s(\xvect)) ^2 \right)
\]

The following proposition is straightforward:
\begin{proposition}
If the interest points of an image $I$ are detected with the
cornerness measurement ${\cal C}$ and with a threshold $t$ such that
${\cal C}>t$, then at scale $s$ the interest points are detected with
$s^4{\cal C}_s>t$.
\end{proposition}

\begin{figure*}[p]
\centerline{
\resizebox{\textwidth}{!}{\begin{tabular}{c@{\hspace{2mm}}cc}
\includegraphics[width=0.20\textwidth]{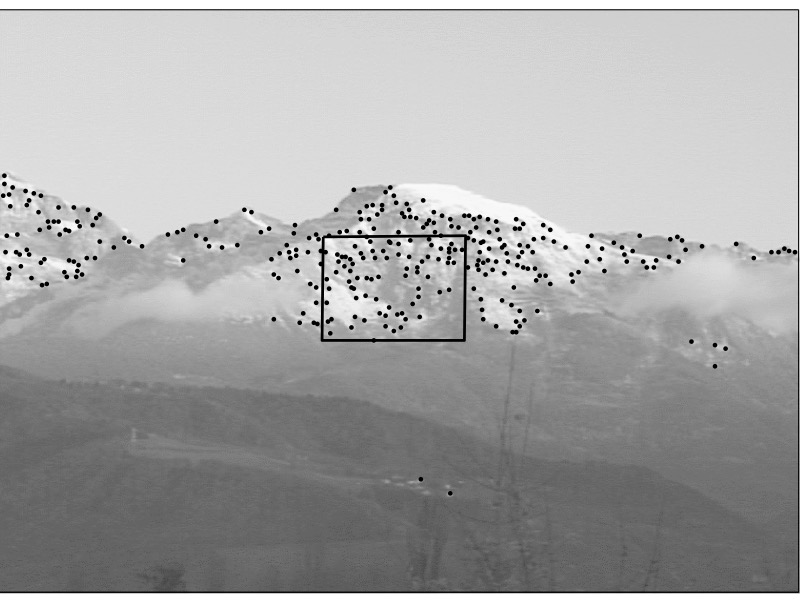}&
\includegraphics[width=0.2\textwidth]{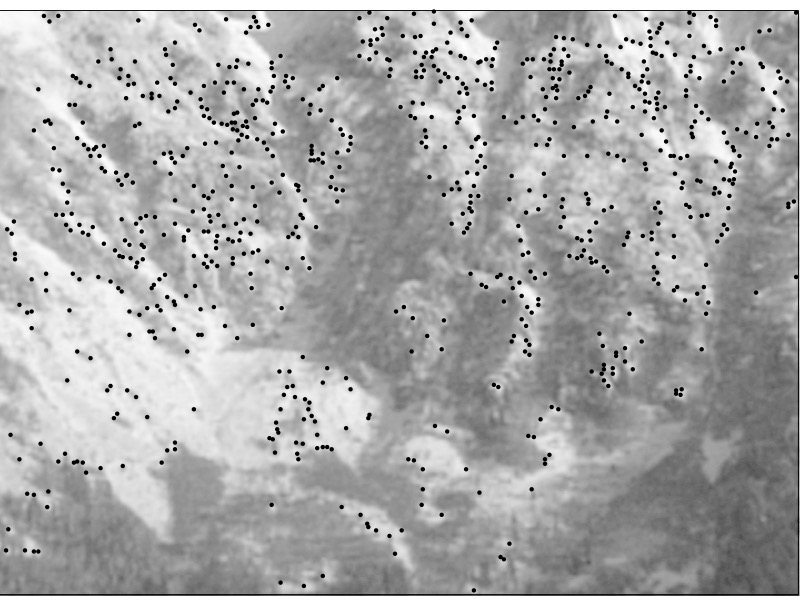}&$s=1$\\
\includegraphics[width=0.2\textwidth]{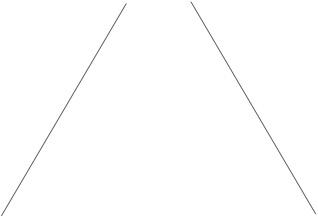}&
\includegraphics[width=0.2\textwidth]{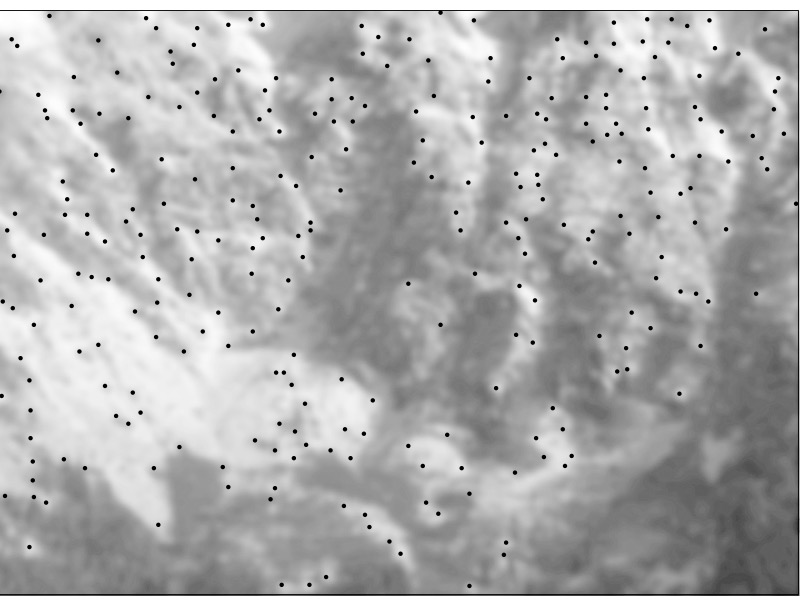}&$s=3$\\
\includegraphics[width=0.2\textwidth]{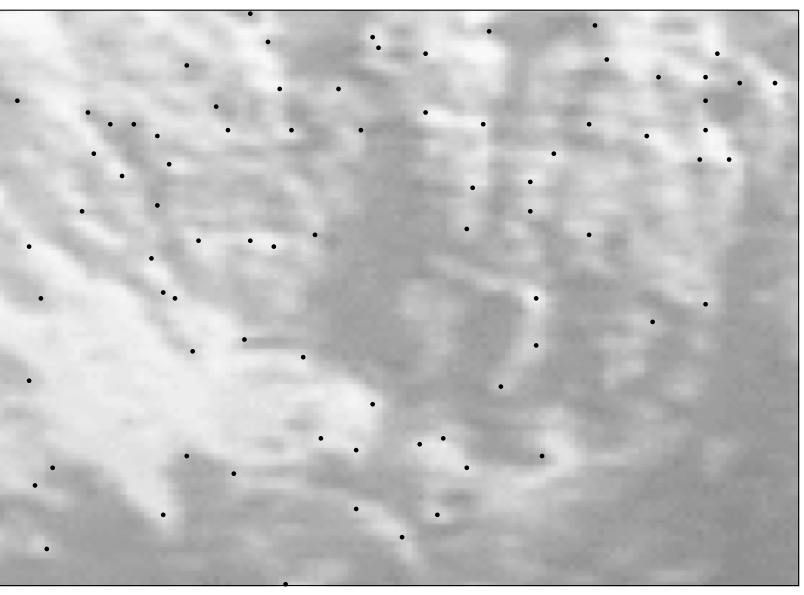}&
\includegraphics[width=0.2\textwidth]{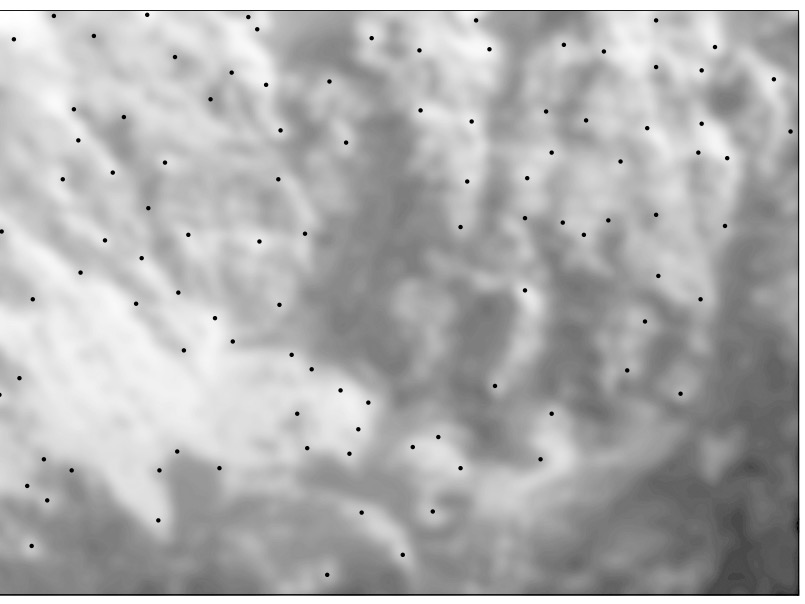}&$s=5$\\
&
\includegraphics[width=0.2\textwidth]{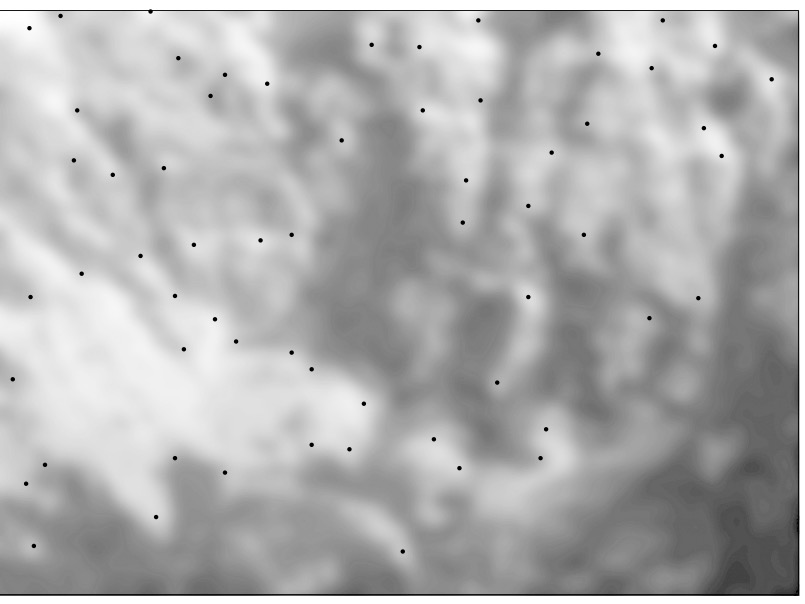}&$s=7$\\
\end{tabular}
}}
\caption{Interest points detected at 4 scales (left) and the points
detected in the corresponding low-resolution image (right).}
\label{fig:Harris-8scales}
\end{figure*}

In order to illustrate the results obtained with this scale-space
interest-point detector, we applied it to the high-resolution image of 
Figure~\ref{fig:Grand-Colon-image-pair} (right).
Figure~\ref{fig:Harris-8scales} shows these results with $\sigma=1$ and
$\tilde{\sigma}=2$. The left side of this figure shows the interest
points detected in the low-resolution image. The image region
corresponding to the high resolution image is zoomed out by a factor of
$5.3$ which is the true scale factor between the two images. The
right side of this figure shows the high-resolution image with
interest points detected at four different scales, 1, 3, 5, and 7. The
best matching scale is shown, side by side, with the zoomed-out
low-resolution region. This is clear evidence that the scale-space
representation and detection of interest points facilitates the
matching task.

\begin{figure}[t!]
\begin{center}
\includegraphics[width=0.95\columnwidth]{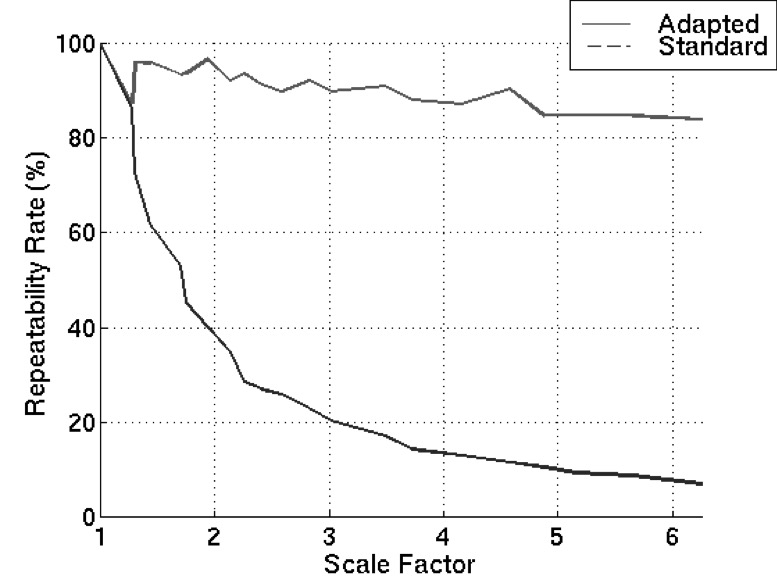}
\end{center}
\caption{Comparison of the standard Harris detector ({\it Standard} or 
bottom curve)
and the scale-space version ({\it Adapted} or top curve). The comparison criteria
is the repeatability rate which is displayed as a function of the scale
factor.} 
\label{fig:repeatability}
\end{figure}

The importance of adapting the scale for interest
point description and detection is shown on
Figure~\ref{fig:repeatability}. This figure shows a comparison between 
the standard Harris
detector and the scale-space interest point detector. The scale
factor varies
from 1 to 6. The scale-space version uses the known
scale factor between test images to adapt the interest point
detection. The measure used in order to
evaluate the performance is the repeatability rate introduced and
thoroughly investigated in~\cite{schmid00a}.
This measure takes into account the number of points repeated between the reference image and
the scaled image with respect to the total number of points. One may
clearly see that the scale-space detector shows very good performance. In
the case of the standard detector the results are insufficient above a
scale factor of 2 (less than 40\% of the points are repeated). 

\section{Robust image matching}

The scale-space extraction and representation of interest points 
will enable us to devise an
image matching method. The main idea is 
to compare the low-resolution image at one scale with the
high-resolution image at many scales. The scale at which the matching
performs the best corresponds to the largest set of point-to-point
assignments between a low-resolution image region and the
high-resolution image.

Without loss of generality, while the low-resolution image $I'$ is
represented at one scale, the high-resolution image $I$
is represented at 8 different scales $\sigma$, 2$\sigma$, \ldots,
8$\sigma$ with $\sigma=1$. At each scale $s_i$, interest points are
extracted using eq.~(\ref{eq:Harris-scale}). Furthermore, each
interest point (in both images and at all scales) is characterized by a
description-vector whose elements are differential invariants. These invariants
were introduced by Koenderink et al. \cite{konderink87a} and were
adapted for image matching by Schmid \& Mohr \cite{schmid97a}.

Following Schmid \& Mohr \cite{schmid97a} two points of interest
match if the Mahalanobis distance between their associated descriptors 
is small. Let $\Vvect_m$ be a description-vector associated with point 
$m$. The distance between two points, $m$ and $m'$ writes:
\begin{equation}
d_{\cal M}(m,m') = \sqrt{(\Vvect_m - \Vvect_{m'})\tp
\mm{\Lambda}\inverse(\Vvect_m - \Vvect_{m'})}
\label{eq:mahalanobis}
\end{equation}

This distance selects potentially good matches but is not powerful
enough because it does not take into account neither local
configurations of image points nor the global geometric transformation
between the two images.

\subsection{Matching based on local collections of points}

One way to disambiguate point matches is to consider collections of
interest points in a small image region and to try to match mutually
compatible sets of points rather than individual points. Here
compatibility is understood both in the sense of topology and
geometry. The concept of mutually compatible feature matches stems
from earlier work in 2-D object recognition \cite{BollesCain82}, 3-D
object recognition \cite{BollesHoraud86}, \cite{FaugerasHebert86}, and 
stereo matching \cite{HoraudSkordas89}. 

Here we are interested in considering a match $(m-m')$, a
neighbourhood $N(m)$ around point $m$, and a neighbourhood  $N(m')$ around
$m'$. We seek to establish whether there are other point
matches within these two neighbourhoods which are topologically,
photometrically, and geometrically compatible. Let $k$ be the number
of point matches based on the Mahalanobis distance:
$(m_1-m'_1)$, \ldots $(m_j-m'_j)$, \ldots $(m_k-m'_k)$, such that
$m_j\in N(m)$, $m_j\neq m$ and $m'_j\in N(m')$, $m'_j\neq m'$ for all $j$, $1\leq j\leq k$.

\begin{figure*}[t!]
\centerline{
\resizebox{0.8\textwidth}{!}{\begin{tabular}{c@{\hspace{4mm}}c}
\includegraphics[width=\textwidth]{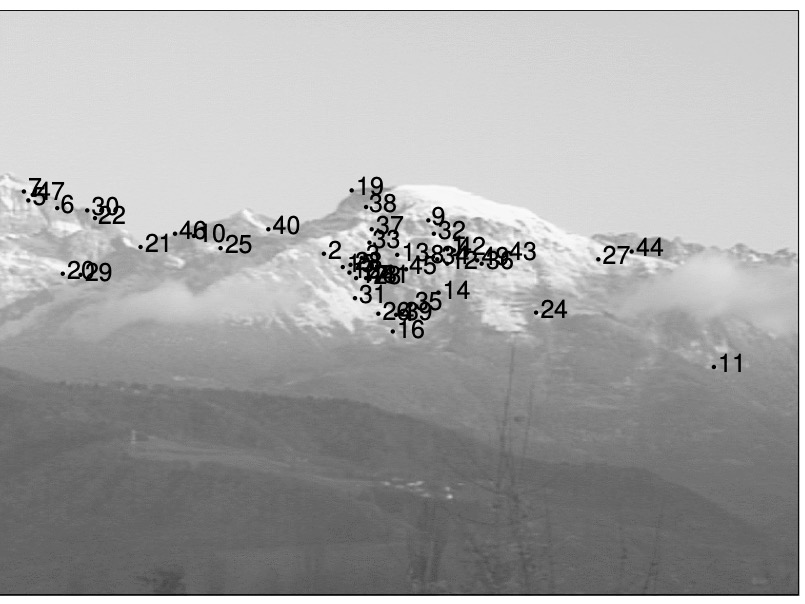}&
\includegraphics[width=\textwidth]{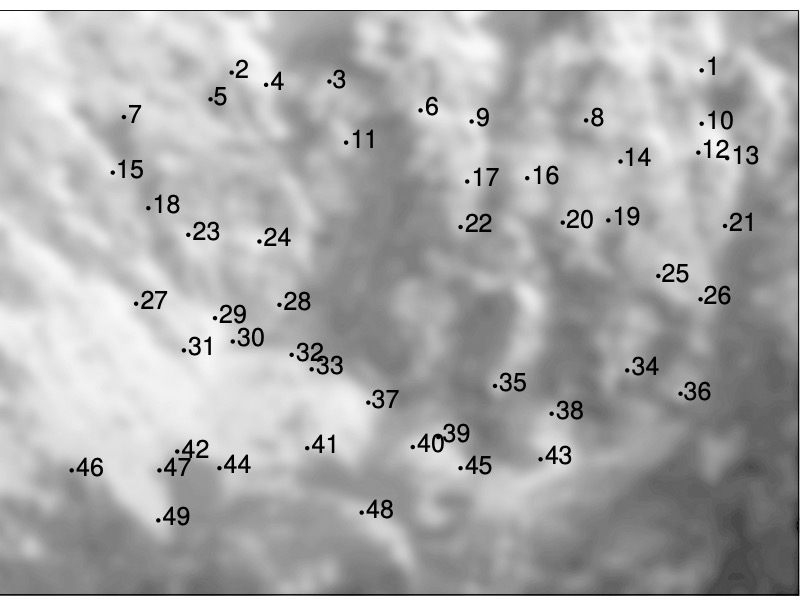}
\end{tabular}
}}
\caption{Matching points using the Mahalanobis distance between their
description vectors.}
\label{fig:point-matching}
\end{figure*}

\begin{figure*}[h!]
\centerline{
\resizebox{0.8\textwidth}{!}{\begin{tabular}{c@{\hspace{4mm}}c}
\includegraphics[width=1.\textwidth]{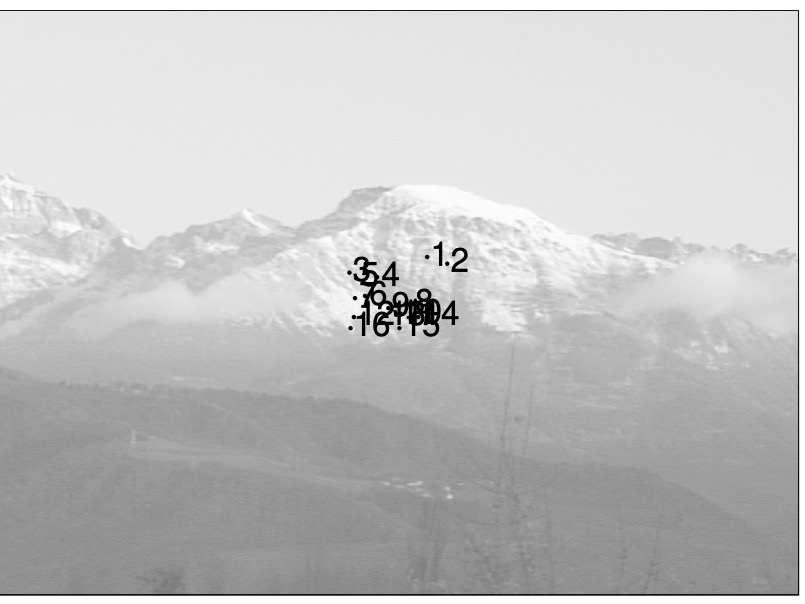}&
\includegraphics[width=1.\textwidth]{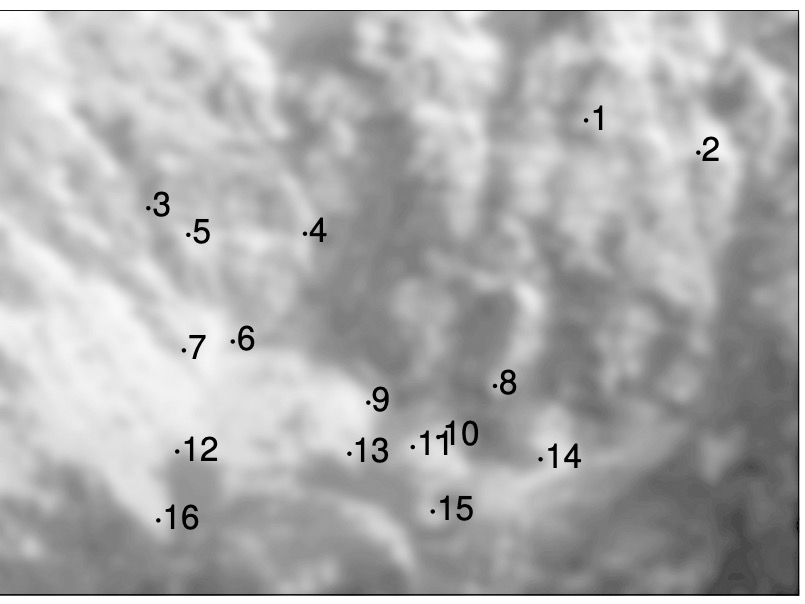}
\end{tabular}
}}
\caption{Matching points using constraints based on local collections
of points.}
\label{fig:point-constraints}
\end{figure*}

These point-to-point matches allow to compute a similarity
transformation between the two regions along the following lines:
\begin{enumerate}
\item select two matches (the central match plus an additional one), 
\item compute 
the parameters of the associated similarity transformation,
e.g. eq.~(\ref{eq:plane-affinity}), 
\item verify how many other matches in
the neighbourhood are consistent with these parameters, 
\item etc. 
\end{enumerate}
This matching method is
implemented as a depth-first tree search. A final test based on
eq.~(\ref{eq:similarity-measure}) allows to assess the match.
The difference between matching points without local support and with
local support is illustrated on figures \ref{fig:point-matching} and
\ref{fig:point-constraints}. The image shown onto the left is the
low-resolution image. The image shown onto the right is the high
resolution image which is represented here at scale $5.3\sigma$ -- the 
true scale factor between the two images. Figure
\ref{fig:point-matching} shows point matches established based on the
Mahalanobis distance while Figure \ref{fig:point-constraints} shows
the result of matching using the method just described. 

\begin{figure*}[h!]
\centerline{
\resizebox{0.8\textwidth}{!}{\begin{tabular}{c@{\hspace{4mm}}c}
\includegraphics[width=1.\textwidth]{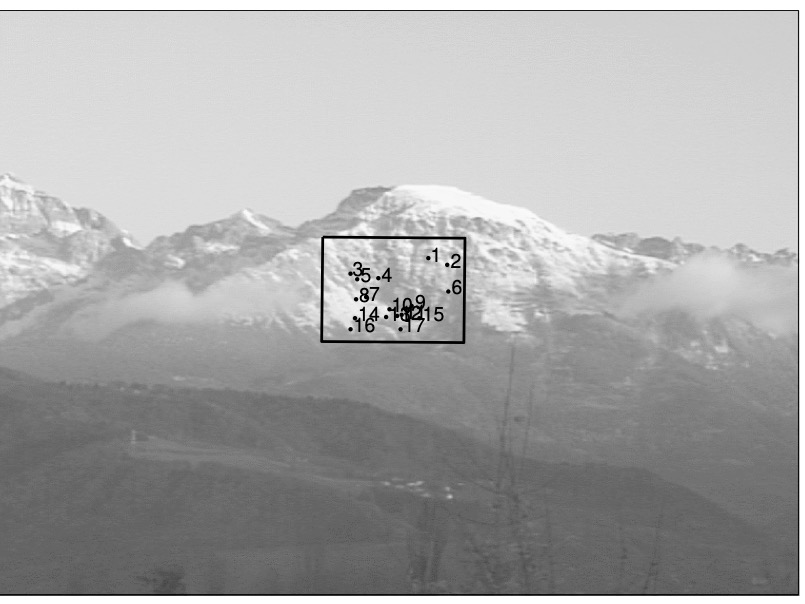} &
\includegraphics[width=1.\textwidth]{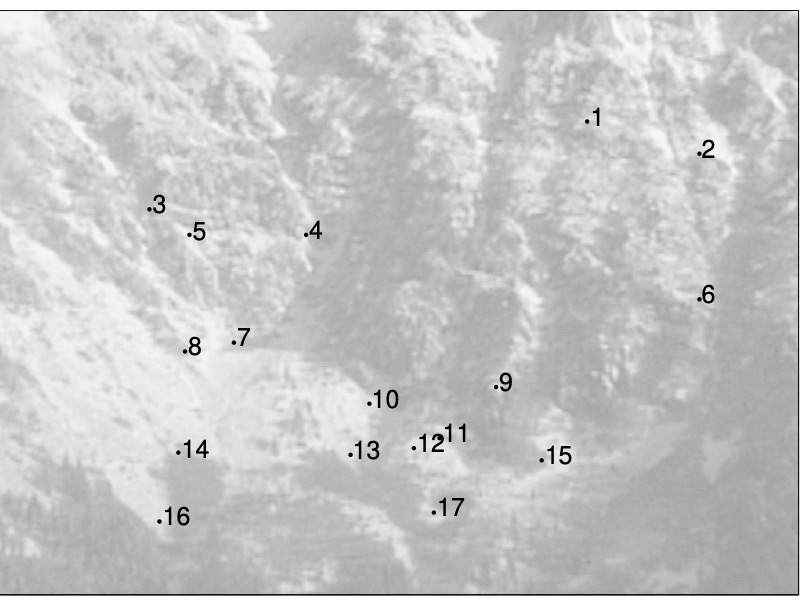}
\end{tabular}
}}
\caption{Matching result for the image pair in Figure
\protect{\ref{fig:Grand-Colon-image-pair}}.
The high-resolution image
is mapped onto the low-resolution one using the similarity estimated from
the 17 matches.} 
\label{fig:result-Grand-Colon}
\end{figure*}

\subsection{Matching at different scales}

The matching algorithm considers, one by one, the scale-space
representations of the high resolution image
and attempts to find which one of these images best matches a
region in the low resolution image. Since there is a strong relationship between
scale and resolution, one may assume that the scale of the
best match roughly corresponds to the resolution ratio between the two 
images. The final exact transformation between image and region 
is found by estimating the associated similarity.

Once an approximate scale has been selected using this strategy, a
robust estimator takes as input the potential one-to-one 
point assignments, computes the best transformation between 
the two images, and splits the point assignments into two sets:
(1)~inliers, i.e., points lying within the small region corresponding to the
similartity mapping of the high resolution image onto the low resolution
one and (2)~outliers, i.e., points that are either outside this region 
or mismatched points inside the region.

Commonly used robust estimators include M-estimators,
least-median-squares (LMedS), and RANdom SAmple Consensus
(RANSAC). In our case, the number of outliers may be quite large. This 
occurs in particular when the two images have very different
resolutions and hence only 20\% or less of the low-resolution image
corresponds to the high resolution one. Therefore, we ruled out
M-estimators because they tolerate only a few outliers. Among the two
remaining techniques, we preferred RANSAC because it allows the user
to define in advance the number of potential outliers through the
selection of a threshold. Hence, this threshold can be chosen as a
function of the scale factor. Details concerning threshold selection
can be found in \cite{CsurkaDemirdjianHoraud99}.

\section{Experiments}

\begin{figure*}[p]
\centerline{
\resizebox{0.8\textwidth}{!}{\begin{tabular}{c@{\hspace{2mm}}cc}
\includegraphics[width=0.25\textwidth]{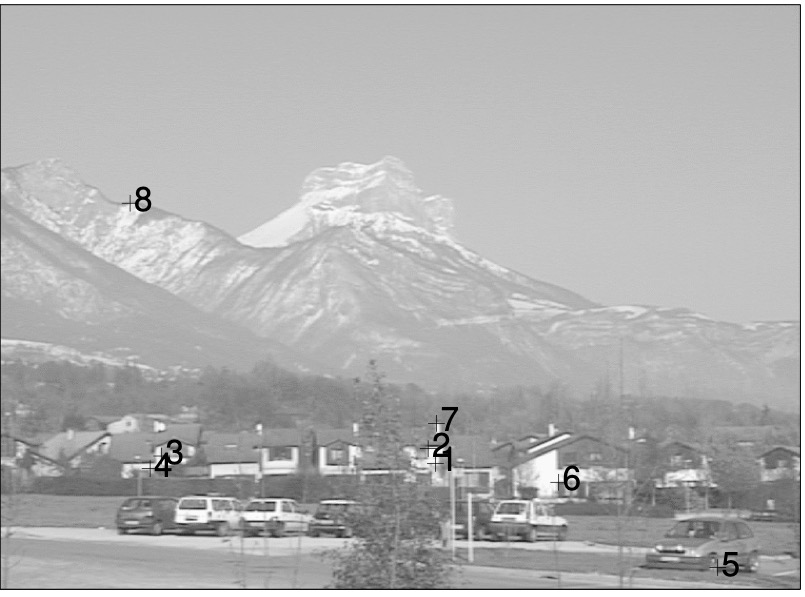} &
\includegraphics[width=0.25\textwidth]{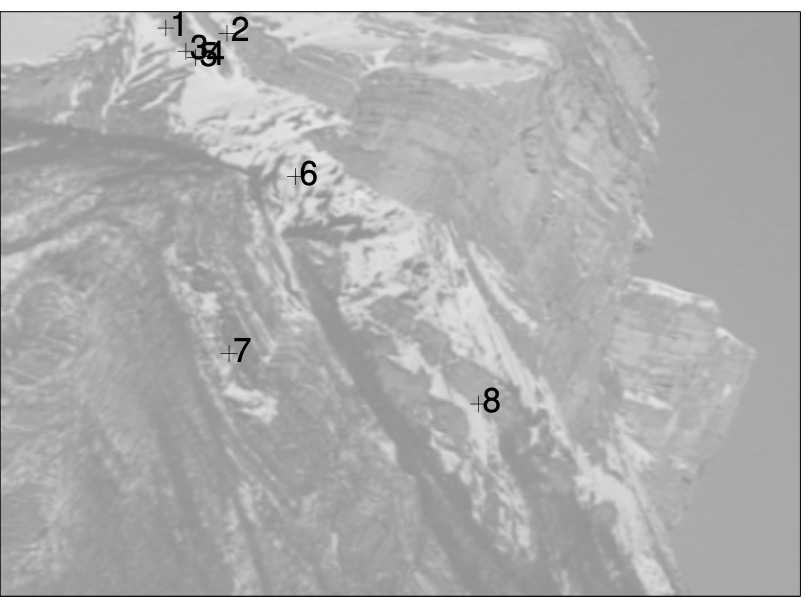} \\
\includegraphics[width=0.25\textwidth]{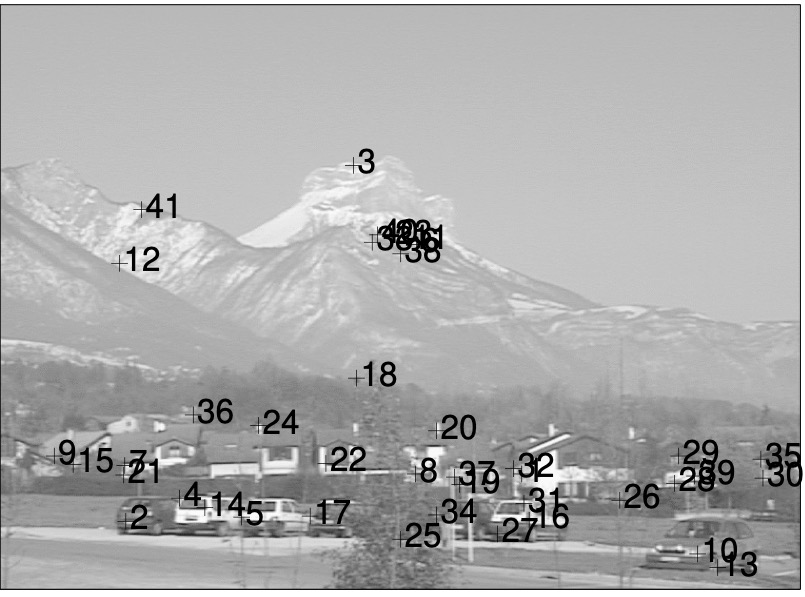} &
\includegraphics[width=0.25\textwidth]{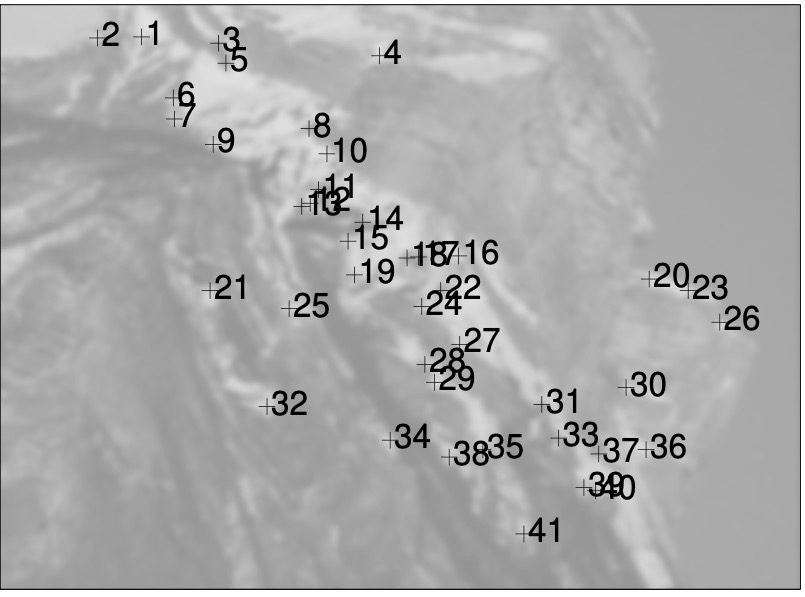} \\
\includegraphics[width=0.25\textwidth]{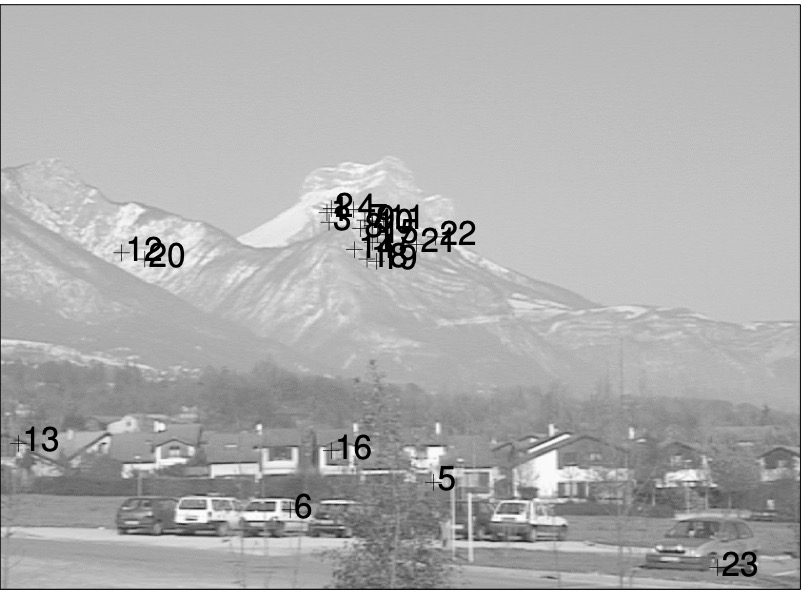} &
\includegraphics[width=0.25\textwidth]{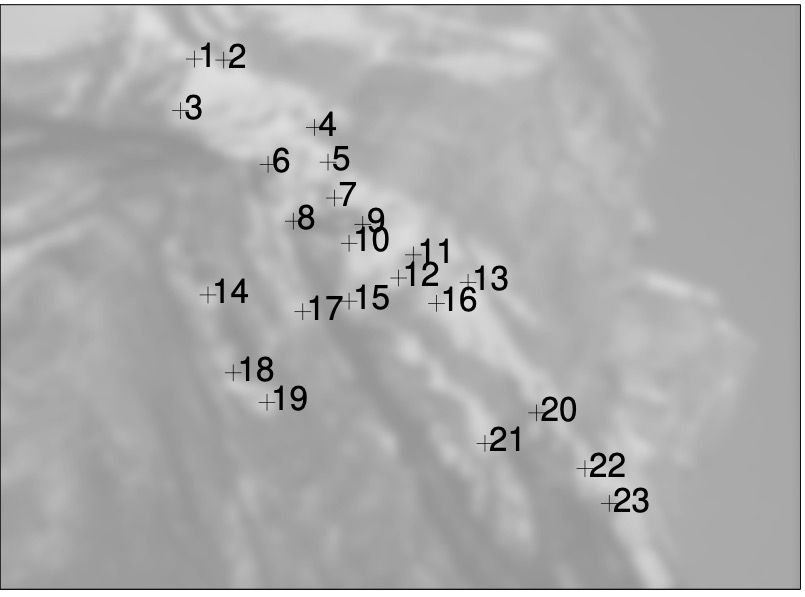} \\
\includegraphics[width=0.25\textwidth]{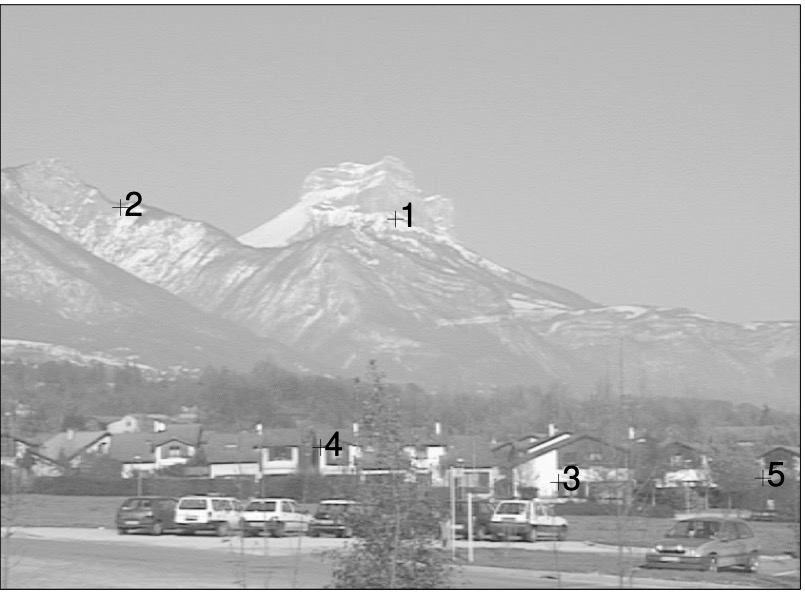} &
\includegraphics[width=0.25\textwidth]{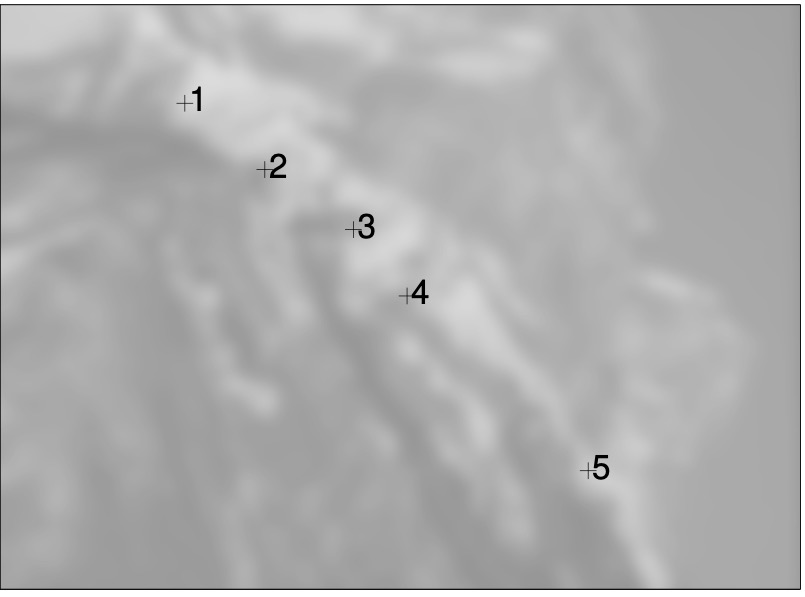}
\end{tabular}
}}
\caption{Initial point-to-point assignments obtained at
four scales (1,3,5,8). The true resolution factor between the two images is 5.}
\label{fig:match-crolles}
\end{figure*}

\begin{figure*}[p]
\centerline{
\resizebox{0.8\textwidth}{!}{\begin{tabular}{c@{\hspace{2mm}}cc}
\includegraphics[width=0.25\textwidth]{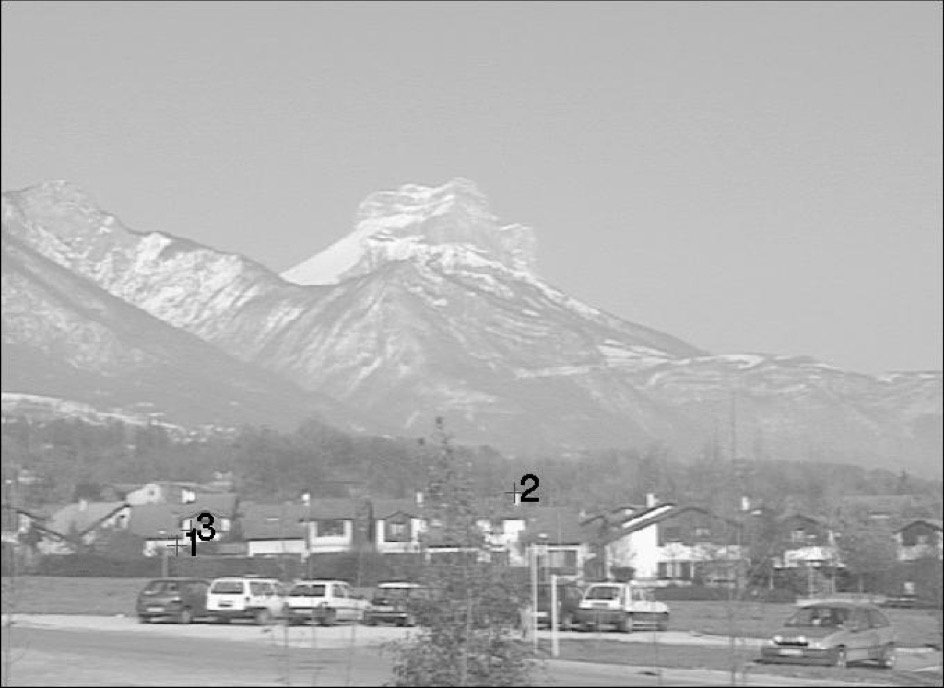} &
\includegraphics[width=0.25\textwidth]{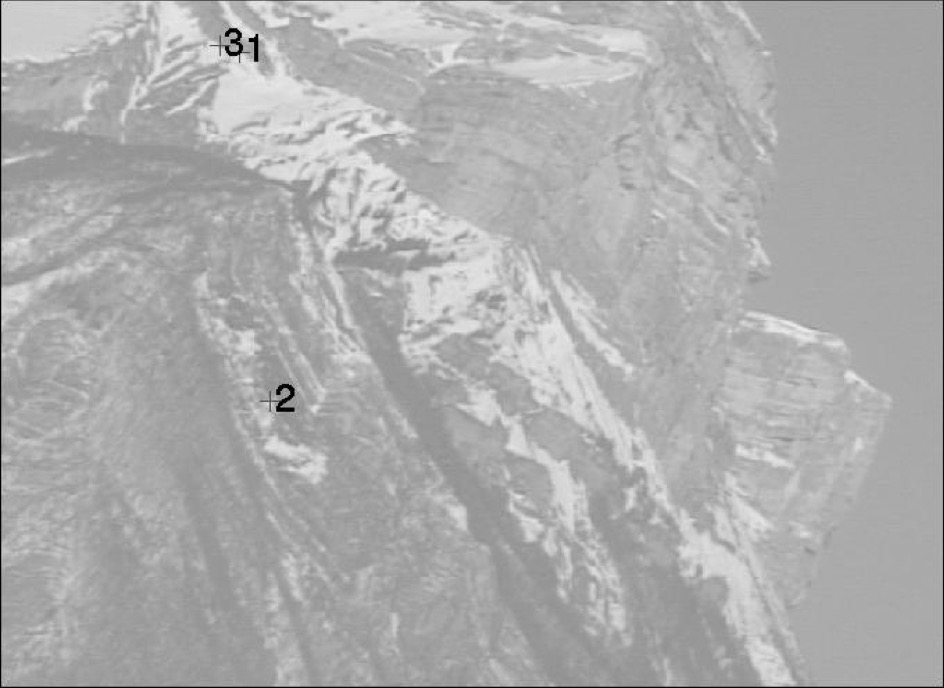} \\
\includegraphics[width=0.25\textwidth]{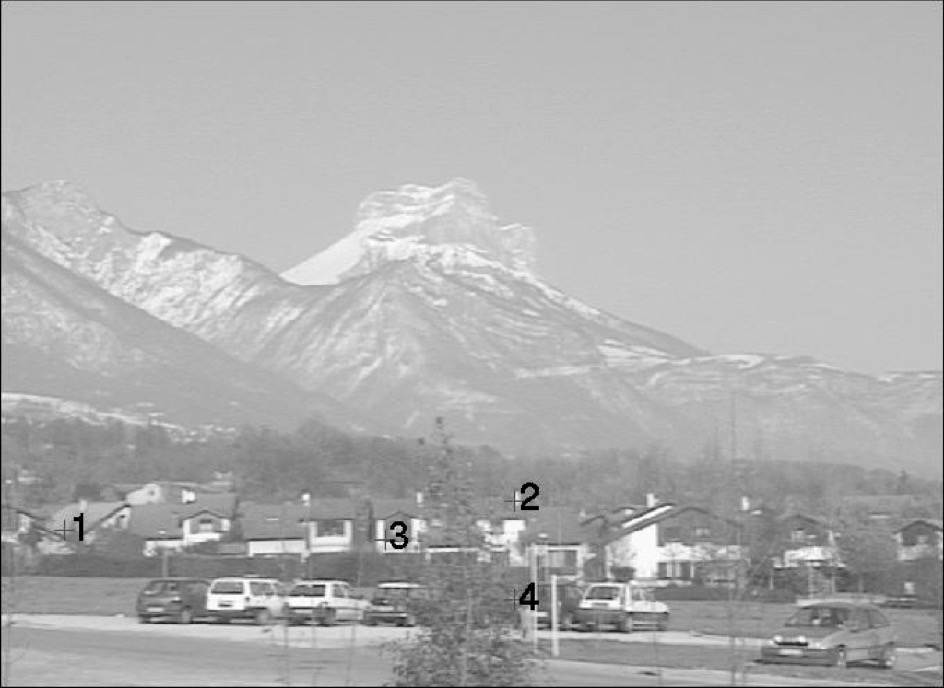} &
\includegraphics[width=0.25\textwidth]{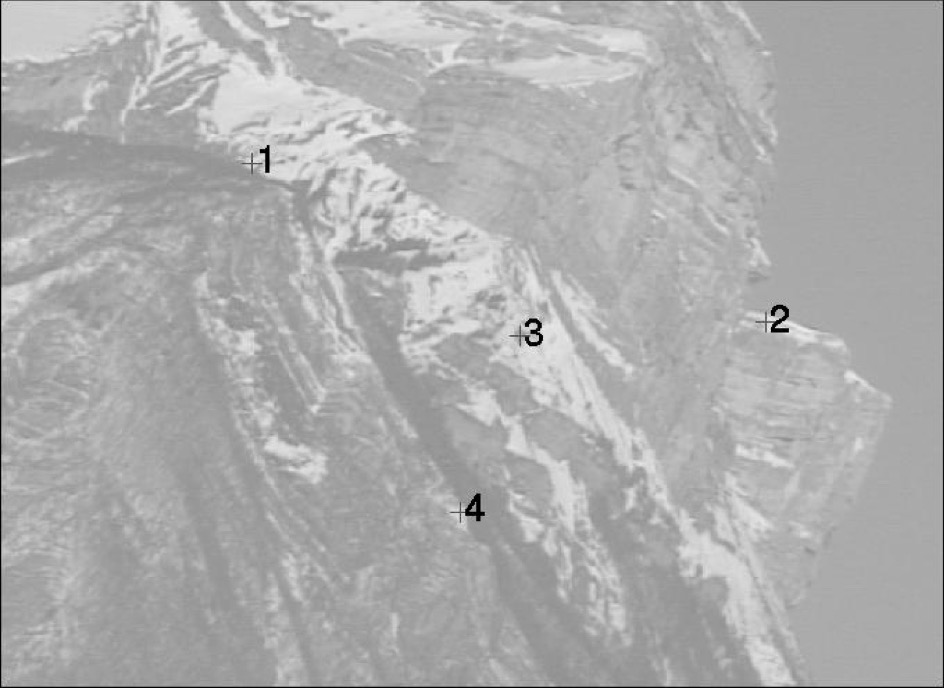} \\
\includegraphics[width=0.25\textwidth]{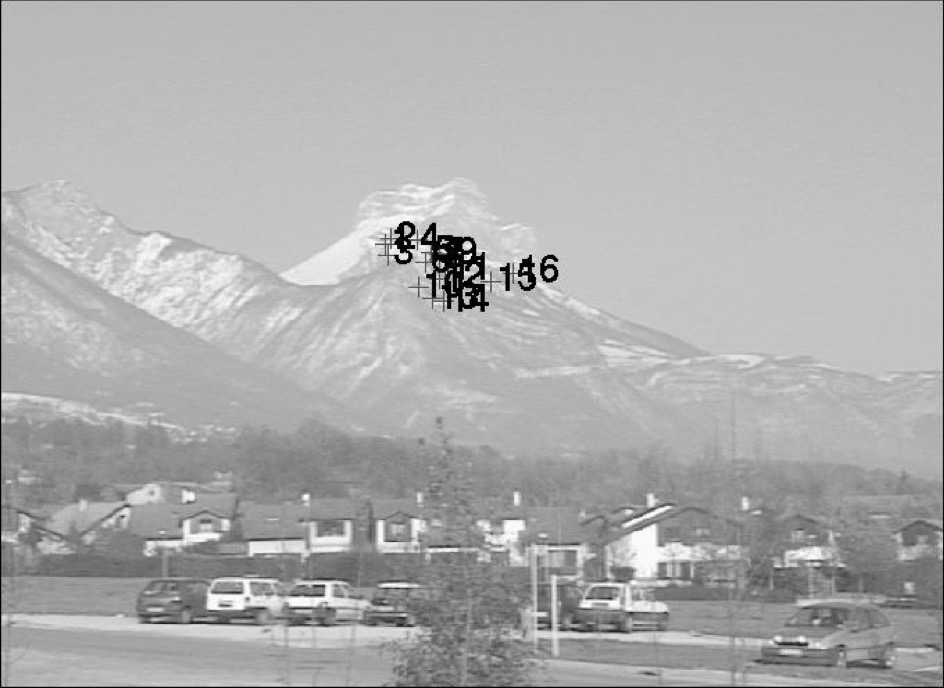} &
\includegraphics[width=0.25\textwidth]{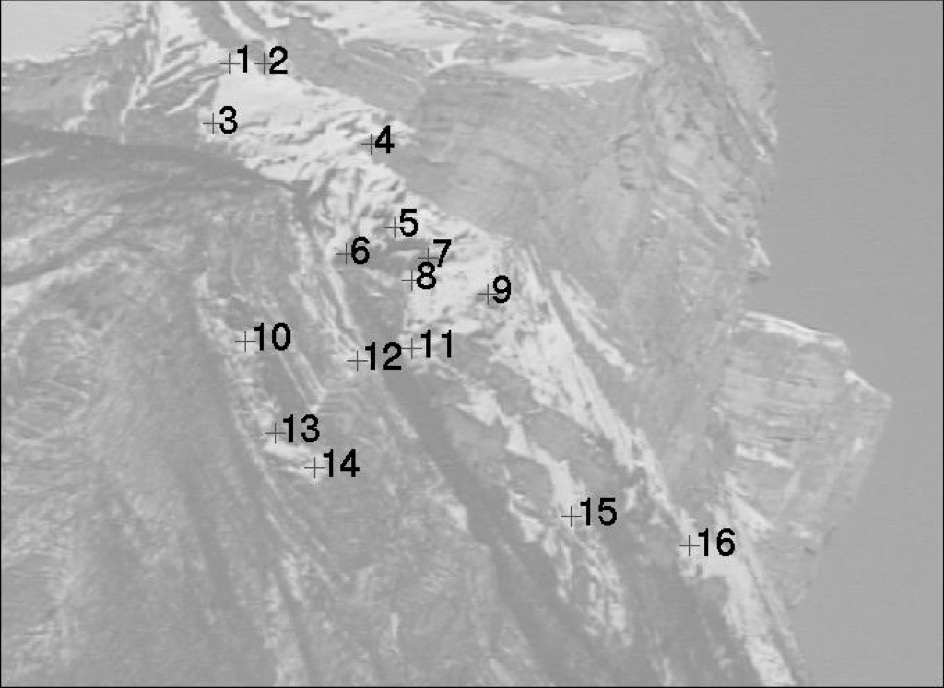} \\
\includegraphics[width=0.25\textwidth]{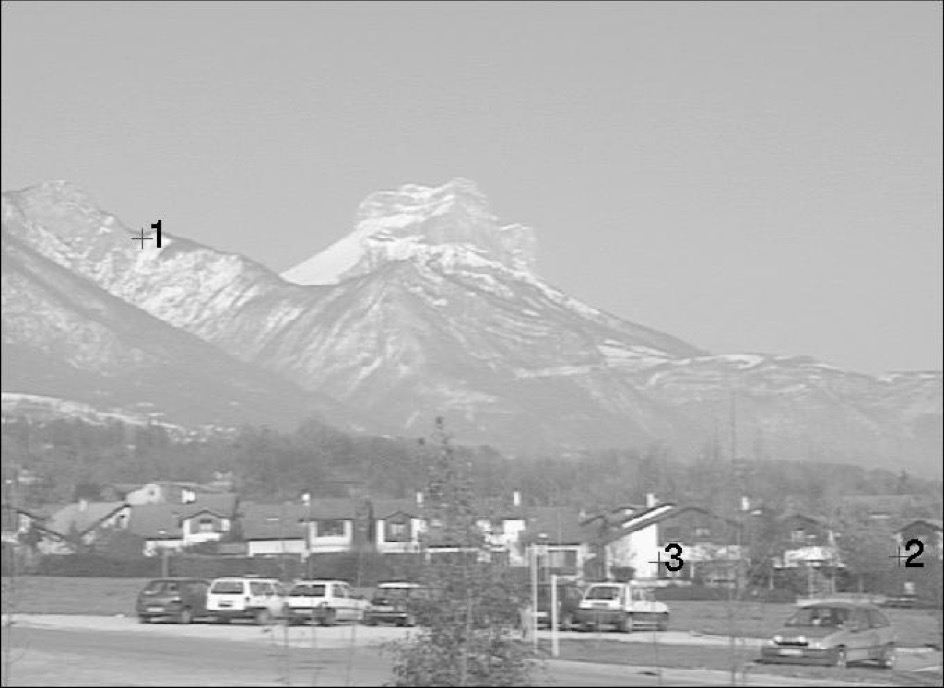} &
\includegraphics[width=0.25\textwidth]{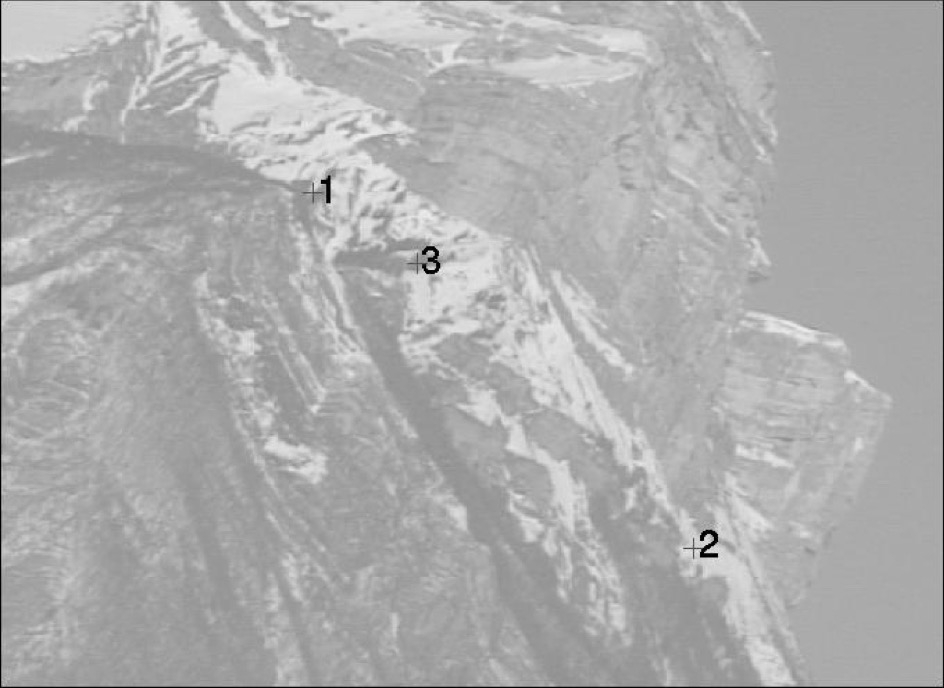}
\end{tabular}
}}
\caption{Inliers after applying the local constraints and the robust
estimator to the previous results.}
\label{fig:inliers-crolles}
\end{figure*}

\begin{figure*}[h!t!]
\centerline{
\includegraphics[width=0.4\textwidth]{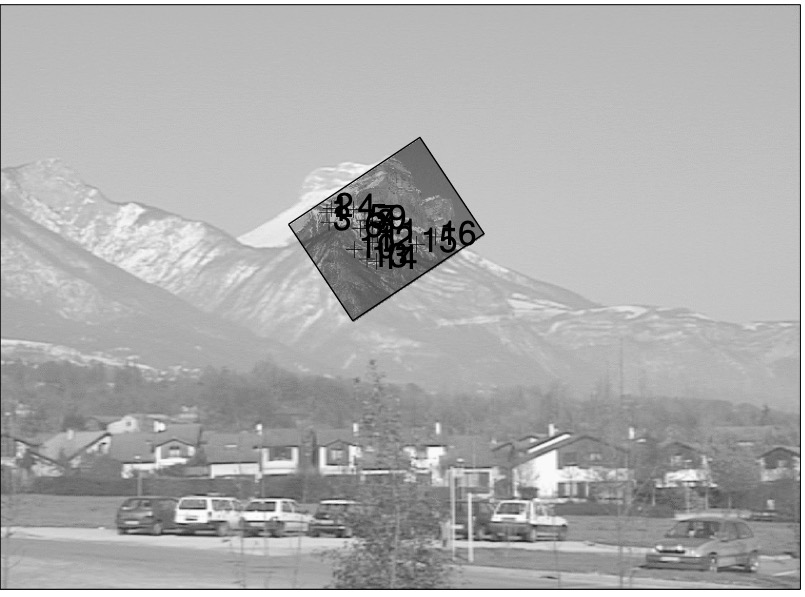} 
\includegraphics[width=0.4\textwidth]{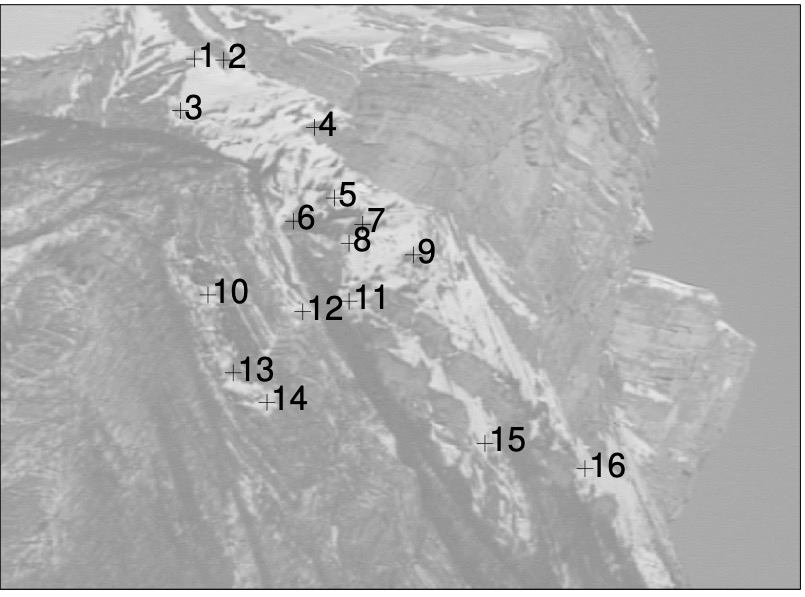}
}
\caption{The final result obtained for the example in Figure
\protect{\ref{fig:match-crolles}}.
All of the 16 matches are correct. The high-resolution image
is mapped onto the low-resolution one using the homography consistent
with the 16 matches. The estimated  rotation angle is 34 degrees and
the estimated resolution change is 5.} 
\label{fig:result-crolles}
\end{figure*}

The matching strategy just described was applied and tested over a
large number of image pairs where the resolution factor between the
two images varied from 2 to 6. 
Let us explain in detail how this type of result is obtained for
another example, e.g.,
Table~1 and
Figures~\ref{fig:match-crolles}, \ref{fig:inliers-crolles}, and
\ref{fig:result-crolles}. Interest 
points are first extracted from the low-resolution image at one 
scale ($s=1$) and from the high-resolution image at 8 different scales
($1$ to $8$). Therefore, eight image matchings are performed. 
Figure~\ref{fig:match-crolles} shows the results
of the  point-to-point matching based on the Mahalanobis distance at
four different scales: 1, 3, 5, and 
8. These results correspond to the column named ``Initial'' in Table~1.
Obviously, scales 3 and 5 have the best matches associated with
them and scale 5 is a better candidate. Therefore, it would have been
sufficient to run the remainder of the  matching algorithm at scale 5 only. In
practice we run the latter algorithm at all
scales. 

These  initial matches are used for enforcing the local constraints
and for
the robust estimation of the
similarity transformation. Figure~\ref{fig:inliers-crolles}
shows the results of applying both these two stages of the algorithm. The
results are summarized in Table~1 in the column ``Inliers''.
One may verify that the best match is obtained at $s=5$. Out of 25
points detected at this scale, 23 among them have a 
potential assignment in the low-resolution image and 16 among them are
finally selected by the robust matching technique. The latter rejected
30\% of the matches. Notice that the resolution factor computed from the
homography is correct for $s=4$, $s=5$ and $s=6$.  
Finally the image-to-region transformation thus obtained was applied to the high
resolution image and this image is reproduced on top of the
low-resolution one (cf. Figure~\ref{fig:result-crolles}). 

\begin{figure*}[htb]
\begin{center}
\includegraphics[width=0.4\textwidth]{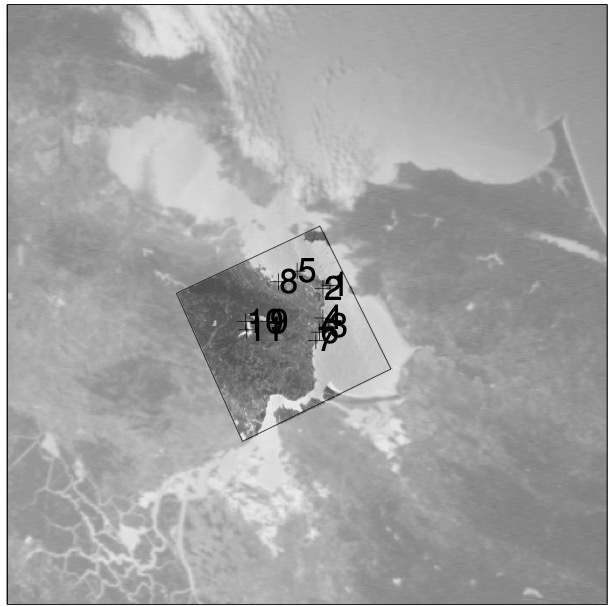}
\includegraphics[width=0.4\textwidth]{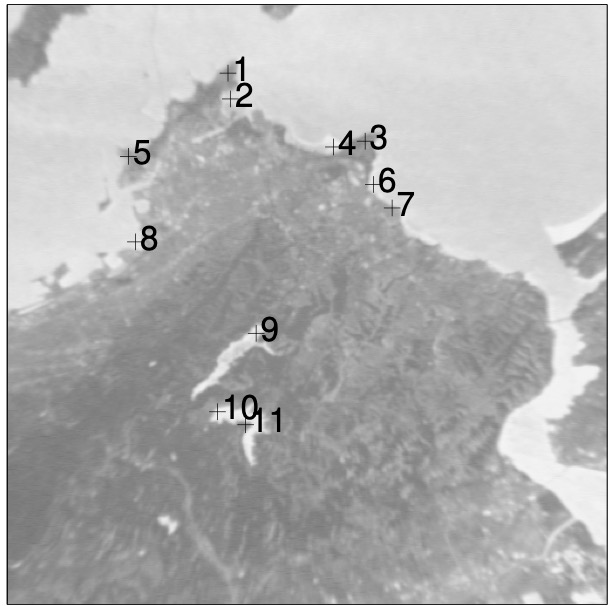}
\end{center}
\caption{Example for a 2D scene. All of the 11 matches are
correct. The estimated rotation angle is 65 degrees and the estimated
resolution change 3.7.}
\label{fig:earth:match}
\end{figure*}

\begin{figure*}[htb]
\begin{center}
\includegraphics[width=0.4\textwidth]{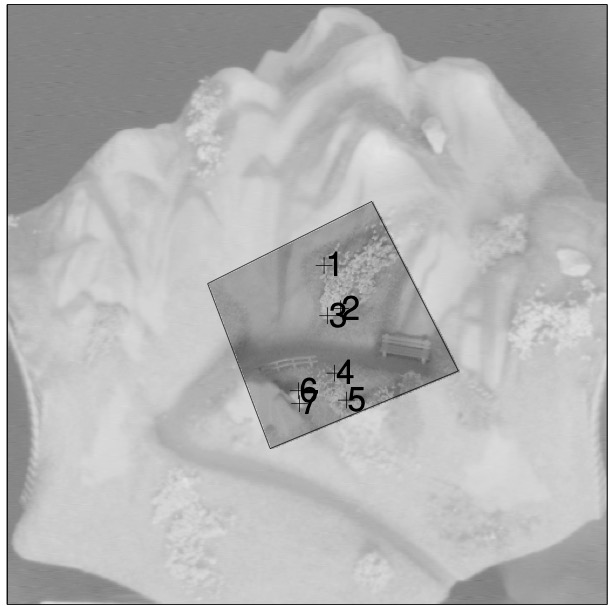}
\includegraphics[width=0.4\textwidth]{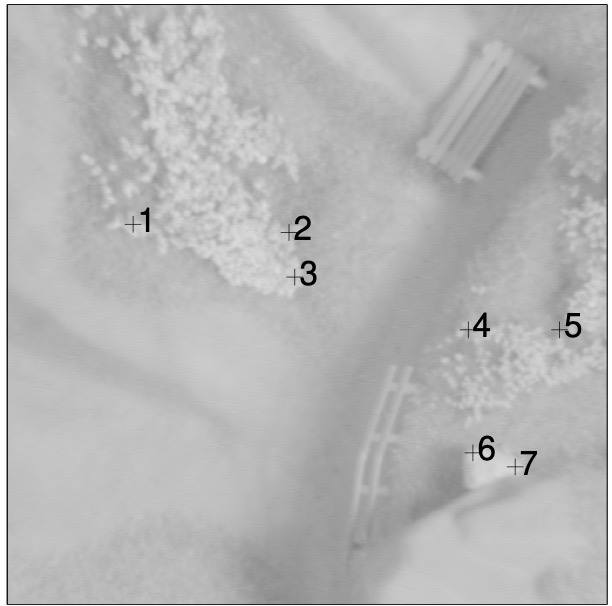}
\end{center}
\caption{Example for the 3D scene ``tunnel''.  All of the 7 matches
are correct. The estimated rotation angle is 77 degrees and the
estimated resolution change is 3.2.} 
\label{fig:tunnel:match}
\end{figure*}

\begin{figure*}[htb]
\begin{center}
\includegraphics[width=0.4\textwidth]{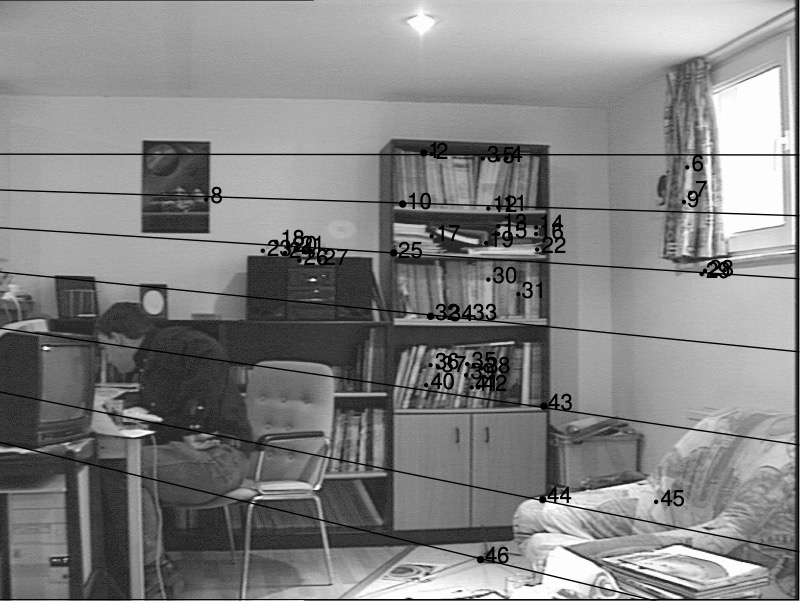}
\includegraphics[width=0.4\textwidth]{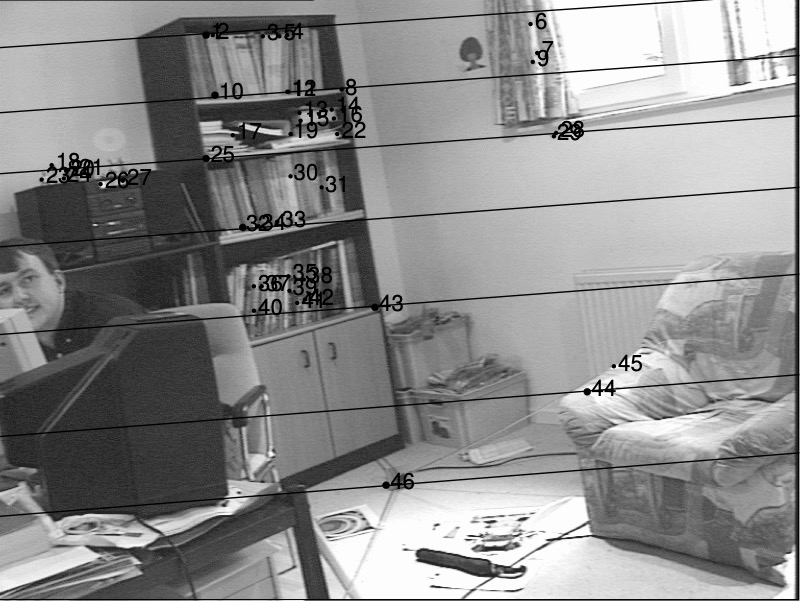}
\end{center}
\caption{This figure shows the epipolar geometry as computed with the
matching and estimation method described in this paper. Notice the
large discrepancy in the viewpoints associated with the two
images. The matcher seems to give advantage to collections of coplanar 
points.} 
\label{fig:example-epipolar}
\end{figure*}

\begin{table}
\begin{center}
\begin{tabular}{|c|c||c|c|c|c|}
\hline
\multicolumn{2}{|c||}{{\small Scale factor}} &
\multicolumn{1}{|c|}{ {\small N$^o$ of points} } &
\multicolumn{3}{|c|}{ {\small N$^o$ matches}}\\
\hline
$s$ & {\small estimated}  && {\small initial}  & {\small
inliers} & {\small  outliers} \\
\hline
1 &  1.3 &    329 &     8 &    - &   100 \%   \\
2 &  0.7 &    126 &    64 &    4 &   94 \% \\
3 &  1.8 &     64 &    41 &    4 &   90 \% \\
4 &  5   &     31 &    26 &   10 &   62 \% \\
5 &  5   &     25 &    23 &   16 &   30 \% \\
6 &  5   &     18 &    17 &   12 &   29 \% \\
7 &  1.1 &     14 &    14 &    - &      100 \%\\
8 &  0.4 &      5 &     5 &    - &      100 \%\\
\hline
\end{tabular}
\end{center}
\label{tab:Crolles-results}
\caption{This table shows, at each scale, the computed resolution
factor, the number of points in the high-resolution image, the number
of potential matches, the final number of matches, and the percentage
of outliers. Notice that scales 4, 5 and 6 yield very similar results.}
\end{table}

\subsection{Further examples}

So far we have been concerned with matching based on the hypothesis
that there is a similarity transformation between one image and a
region in the other image. This is a relatively restrictive
hypothesis. The following examples show that the matching method
described in this paper may well be applied (with some modifications)
to cases were the two images differ by affine, projective, or epipolar 
transformations. 

The matching strategy remains the same up to the robust estimator. The 
latter uses either an affine transformation, a plane homography, or
the fundamental matrix to confirm matches and to reject outliers. 
Figure~\ref{fig:earth:match} shows an aerial view (left) as well as a
detail (right). An affine transformation was hypothesized and
correctly estimated. A second example (Figure~\ref{fig:tunnel:match})
shows a mock-up, a planar detail, and the correct matches using a plane
homography. 

Figure~\ref{fig:example-epipolar} displays a stereo pair of a complex
3-D scene. The two images are taken from very different viewpoints
with different zoom settings: Classical stereo matching methods
fail to find the epipolar geometry. In spite of some mismatches along
epipolar lines, the epipolar geometry is correctly estimated by the
matching method described in this paper.

\section{Conclusions}
We presented a new method for matching images with
two  very different resolutions. We showed that it is enough to represent
the high-resolution image in scale-space and we described a
one-to-many robust image matching strategy. Key to the success of this
method is the scale-space representation of interest points and their
descriptors. We thoroughly investigated the similarity invariance of
the Harris interest point detector as well as its scale-space
behaviour. Recently this work was extended to characterize the most significant
scale of an interest point and to devise a matching and indexing
method that encapsulates scale changes \cite{mikolajczyk01a}. The extension to affine-invariant local image descriptors
is also on its way \cite{mikolajczyk02a}. 

In spite of a huge number of publications in the image-matching
domain, it seems to us that none of the existing methods is able to
deal with large changes in resolution. Here we have been able to match 
images which differ by a resolution factor up to 6. In practice the images shown in
this paper were gathered by varying the focal length using the zoom-lens of a 
digital camcorder. The advent of digital photography opens new fields
of applications and we believe that our matching technique will allow
the simultaneous exploitation of multiple viewpoints and variable
resolutions.


\appendix
\section{Interest point detection under similarity}
\label{appendix:Harris-over-scale}

In order to prove eq.~(\ref{eq:CandC'}) we consider the convolution
of the Harris operator with a Gaussian kernel, i.e.,
eq.~(\ref{eq:C=GQ}):
\begin{multline*}
\Mmat(\xvect,\sigma,\tilde{\sigma}) = 
G(\xvect,\tilde{\sigma}) \star \Qmat(\xvect, \sigma) \\ = 
\int_U\int_V \Qmat(U,V) G(U-u,V-v,\tilde{\sigma}) dU dV
\end{multline*}
Using eq.~(\ref{eq:QandQ'}) we obtain:
\begin{multline*}
G(\xvect,\tilde{\sigma}) \star \Qmat(\xvect, \sigma) = \\
\int_U\int_V h^2 \Rmat\tp \Qmat'(U',V') \Rmat
G(U-v,V-v,\tilde{\sigma}) dU dV
\end{multline*}
The similarity transformation $\xvect'=h\Rmat\xvect+\tvect$ applied to 
vectors $(U\;V)\tp$ and $(u\;v)\tp$ yields: 
$$dU'dV'=h^2dUdV$$
and
$$(U'-u')^2+(V'-v')^2=h^2((U-u)^2+(V-v)^2)$$ 
Using the latter, the
Gaussian kernel $G(U-u,V-v,\tilde{\sigma})$ becomes:
\begin{align*}
G(U-u,V-v,  \tilde{\sigma}) & = \frac{1}{2\pi\tilde{\sigma}^2} \exp\left(
\frac{(U-u)^2+(V-v)^2}{2\tilde{\sigma}^2} \right) \\
&= h^2
\frac{1}{2\pi(h\tilde{\sigma})^2} \exp\left(
\frac{(U'-u')^2+(V'-v')^2}{2(h\tilde{\sigma})^2} \right)\\
&= h^2 G(U'-u',V'-v', h\tilde{\sigma})
\end{align*}
By substitution we get:
\begin{multline*}
G(\tilde{\sigma}) \star \Qmat(\xvect, \sigma) = \\
h^2 \Rmat\tp \left(
\int_{U'}\int_{V'} \Qmat'(U',V') G(U'-u',V'-v',h\tilde{\sigma}) dU'
dV' \right) \Rmat
\end{multline*}
By taking $\tilde{\sigma}'=h\tilde{\sigma}$ we obtain:
\[
G(\tilde{\sigma}) \star \Qmat(\xvect, \sigma) = h^2 \Rmat\tp \;
G(\tilde{\sigma}') \star \Qmat'(\xvect', \sigma') \; \Rmat
\]
which proves the formula given by eq.~(\ref{eq:CandC'}).

\bibliographystyle{plain}


\end{document}